\documentclass[11pt,a4paper]{article}
\usepackage[hyperref]{acl2019}
\usepackage{times}
\usepackage{latexsym}
\usepackage{algorithmic}
\usepackage{url}
\usepackage{amsmath}
\usepackage{amsfonts}
\usepackage{bm}
\usepackage{booktabs}
\usepackage{tikz}
\usepackage{graphicx}
\usetikzlibrary{bayesnet}
\usepackage{multirow}

\newif{\ifhidecomments}
\hidecommentsfalse

\ifhidecomments
    \newcommand{\nascomment}[1]{}	
    \newcommand{\dbc}[1]{}
    \newcommand{\dallas}[1]{}
    \newcommand{\tam}[1]{}
    \newcommand{\suchin}[1]{}
    \newcommand{\mattp}[1]{}
    \newcommand{\roy}[1]{}
    \newcommand{\pradeep}[1]{}
    \newcommand{\markn}[1]{}
    \newcommand{\gabis}[1]{}
\else
    \newcommand{\nascomment}[1]{\textcolor{blue}{[#1]}}
    \newcommand{\dbc}[1]{\textcolor{orange}{[#1 ---\textsc{DBC}]}}
    \newcommand{\dallas}[1]{\textcolor{orange}{[#1 ---\textsc{DBC}]}}
    \newcommand{\tam}[1]{\textcolor{green}{[#1 ---\textsc{Tam}]}}
    \newcommand{\suchin}[1]{\textcolor{red}{[#1 ---\textsc{SG}]}}
    \newcommand{\mattp}[1]{\textcolor{brown}{[#1 ---\textsc{MP}]}}
    \newcommand{\roy}[1]{\textcolor{purple}{[#1 ---\textsc{RS}]}}
    \newcommand{\pradeep}[1]{\textcolor{violet}{[#1 ---\textsc{PD}]}}
    \newcommand{\markn}[1]{\textcolor{red}{[#1 ---\textsc{MN}]}}
    \newcommand{\gabis}[1]{\textcolor{red}{[#1 ---\textsc{GS}]}}
\fi

\newcommand{\imdb}{{\sc imdb}}
\newcommand{\ag}{{\sc AG}}
\newcommand{\yahoo}{{\sc Yahoo!}}
\newcommand{\hatespeech}{{\sc Hatespeech}}

\newcommand{\elmo}{{\sc ELMo}} 
\newcommand{\bert}{{\sc BERT}}
\newcommand{\ulmfit}{{\sc ULMfit}}
\newcommand{\glove}{{\sc GloVe}}

\usepackage{xspace}

\newcommand{\VAMPIRE}{{\sc vampire}\xspace}

\aclfinalcopy 
\def\aclpaperid{2163} 

\title{Variational Pretraining for Semi-supervised Text Classification}

\newcommand{\andd}{\hspace{1.5em}}
\newcommand{\ai}{$^1$}
\newcommand{\uw}{$^2$}
\newcommand{\cmu}{$^3$}
\author{Suchin Gururangan\ai \andd Tam Dang\uw \andd Dallas Card\cmu \andd Noah A. Smith$^{1,2}$ \vspace{.35em}\\
  \ai Allen Institute for Artificial Intelligence, Seattle, WA, USA  \\
  \uw Paul G. Allen School of Computer Science \& Engineering, University of Washington, Seattle, WA, USA\\ 
  \cmu Machine Learning Department, Carnegie Mellon University, Pittsburgh, PA, USA\\
  {\tt suching@allenai.org \{dangt7,nasmith\}@cs.washington.edu dcard@cmu.edu }
}

\date{}

\begin{document}
\maketitle

\begin{abstract}
We introduce \VAMPIRE,\footnote{VAriational Methods for Pretraining In Resource-limited Environments} a lightweight pretraining framework for effective text classification when data and computing resources are limited.
We pretrain a unigram document model as a variational autoencoder on in-domain, unlabeled data and use its internal states as features in a downstream classifier. Empirically, we show the relative strength of \VAMPIRE~against computationally expensive contextual embeddings and other popular semi-supervised baselines under low resource settings. We also find that fine-tuning to \textit{in-domain} data is crucial to achieving decent performance from contextual embeddings when working with limited supervision. We accompany this paper with code to pretrain and use \VAMPIRE~embeddings in downstream tasks.

\end{abstract}

\section{Introduction}

An effective approach to semi-supervised learning has long been a goal for the NLP community, as unlabeled data tends to be plentiful compared to labeled data.
Early work emphasized using unlabeled data drawn from the same distribution as the labeled data \citep{nigam.2000}, but larger and more reliable gains have been obtained by using contextual embeddings trained with a language modeling (LM) objective on massive amounts of text from domains such as Wikipedia or news \citep{peters.2018,devlin.2018,radford.2018,ruder.2018}.
The latter approaches play to the strengths of high-resource settings (e.g., access to web-scale corpora and powerful machines), but their computational and data requirements can make them less useful in resource-limited environments. 
In this paper, we instead focus on the low-resource setting (\S\ref{sec:environments}), and develop a lightweight approach to pretraining for semi-supervised text classification.

Our model, which we call \VAMPIRE, combines a variational autoencoder (VAE) approach to document modeling \citep{kingma.2013,miao.2016,Srivastava2017AutoencodingVI} with insights from LM pretraining \citep{peters.2018}.
By operating on a bag-of-words representation, we avoid the time complexity and difficulty of training a sequence-to-sequence VAE \citep{bowman.2016,xu.2017,yang.2017} while retaining the freedom to use a multi-layer encoder that can learn useful representations for downstream tasks. Because \VAMPIRE~ignores sequential information, it leads to models that are much cheaper  to train, and offers strong performance when the amount of labeled data is small. Finally, because \VAMPIRE~is a descendant of topic models, we are able to explore model selection by topic \textit{coherence}, rather than validation-set perplexity, which results in better downstream classification performance (\S\ref{sec:stopping}).

In order to evaluate the effectiveness of our method, we experiment with four text classification datasets. We compare our approach to a traditional semi-supervised baseline (self-training), alternative representation learning techniques that have access to the in-domain data, and the full-scale alternative of using large language models trained on out-of-domain data, optionally fine-tuned to the task domain.

Our results demonstrate that effective semi-supervised learning is achievable for  limited-resource settings, without the need for computationally demanding sequence-based  models.
While we observe that fine-tuning a pretrained \bert~model to the domain provides the best results, this depends on the existence of such a model in the relevant language, as well as GPUs to fine-tune it.
When this is not an option, our model offers equivalent or superior performance to the alternatives with minimal computational requirements, especially when working with limited amounts of labeled data.

The major contributions of this paper are:
\begin{itemize}
    \item We adapt variational document models to modern pretraining methods for semi-supervised text classification (\S \ref{sec:models}), and highlight the importance of appropriate criteria for model selection (\S\ref{sec:modelselection}).
    \item We demonstrate experimentally that our method is an efficient and effective approach to semi-supervised text classification when data and computation are limited (\S \ref{sec:results}).
    \item We confirm that fine-tuning is essential when using contextual embeddings for document classification, and provide a summary of practical advice for researchers wishing to use unlabeled data in semi-supervised text classification (\S \ref{sec:discussion}).
    \item We release code to pretrain variational models on unlabeled data and use learned representations in downstream tasks.\footnote{\url{http://github.com/allenai/vampire}}
\end{itemize}

\section{Background}

\subsection{Resource-limited Environments} \label{sec:environments}

In this paper, we are interested in the low-resource setting, which entails limited access to computation, labels, and out-of-domain data. Labeled data can be obtained cheaply for some tasks, but for others, labels may require expensive and time-consuming human annotations, possibly from domain experts, which will limit their availability. 

While there is a huge amount of unlabeled text available for some languages, such as English, this scale of data is not available for all languages. \emph{In-domain} data availability, of course, varies by domain. For many researchers, especially outside of STEM fields, \textit{computation} may also be a scarce resource, such that training contextual embeddings from scratch, or even incorporating them into a model could be prohibitively expensive. 

Moreover, even when such pretrained models are available, they inevitably come with potentially undesirable biases baked in, based on the data on which they were trained \citep{recasens.2013,bolukbasi.2016,zhao.2019}.
Particularly for social science applications, it may be preferable to exclude such confounders by only working with in-domain or curated data.

Given these constraints and limitations, we seek an approach to semi-supervised learning that can leverage in-domain unlabeled data, achieve high accuracy with only a handful of labeled instances, and can run efficiently on a CPU.

\subsection{Semi-supervised Learning}

Many approaches to semi-supervised learning have been developed for NLP, including variants of bootstrapping \citep{charniak.1997,blum.1998,zhou.2005,mcclosky.2006}, and representation learning using generative models or word vectors \citep{mikolov.2013,glove}.
Contextualized embeddings have recently emerged as a powerful way to use out-of-domain data \citep{peters.2018,radford.2018b}, but training these large models requires a massive amount of appropriate data (typically on the order of hundreds of millions of words), and industry-scale computational resources (hundreds of hours on multiple GPUs).\footnote{For example, \ulmfit~was trained on 100 million words, and \bert~used 3.3 billion. While many pretrained models have been made available, they are unlikely to cover every application, especially for rare languages.}

\label{sec:ssmodels}
There have also been attempts to leverage VAEs for semi-supervised learning in NLP, mostly in the form of sequence-to-sequence models \citep{xu.2017,yang.2017}, which use sequence-based encoders and decoders (see \S \ref{sec:models}). These papers report strong performance, but there are many open questions which necessitate further investigation. First, given the reported difficulty of training sequence-to-sequence VAEs \citep{bowman.2016}, it is questionable whether such an approach is useful in practice. Moreover, it is unclear if such complex models (which are expensive to train) are actually required for good performance on tasks such as text classification. 

Here, we instead base our framework on neural document models \cite{miao.2016,Srivastava2017AutoencodingVI,card.2018}, which offer both faster training and an explicit interpretation in the form of \textit{topics}, and explore their utility in the semi-supervised setting.

\section{Model} \label{sec:models}

In this work, we assume 
that we have $L$ documents, $\mathcal{D}_L = \{(\bm{x}_i, y_i)\}_{i=1}^{L}$, with observed categorical labels $y \in \mathcal{Y}$. We also assume access to a larger set of $U$
documents drawn from the same distribution, but for which the labels are unobserved, i.e, $\mathcal{D}_U = \{\bm{x}_i\}_{i=L+1}^{U+L}$. Our primary goal is to learn a probabilistic classifier, $p(y \mid \bm{x})$. 

\begin{figure}[th]
    \includegraphics[scale=0.25]{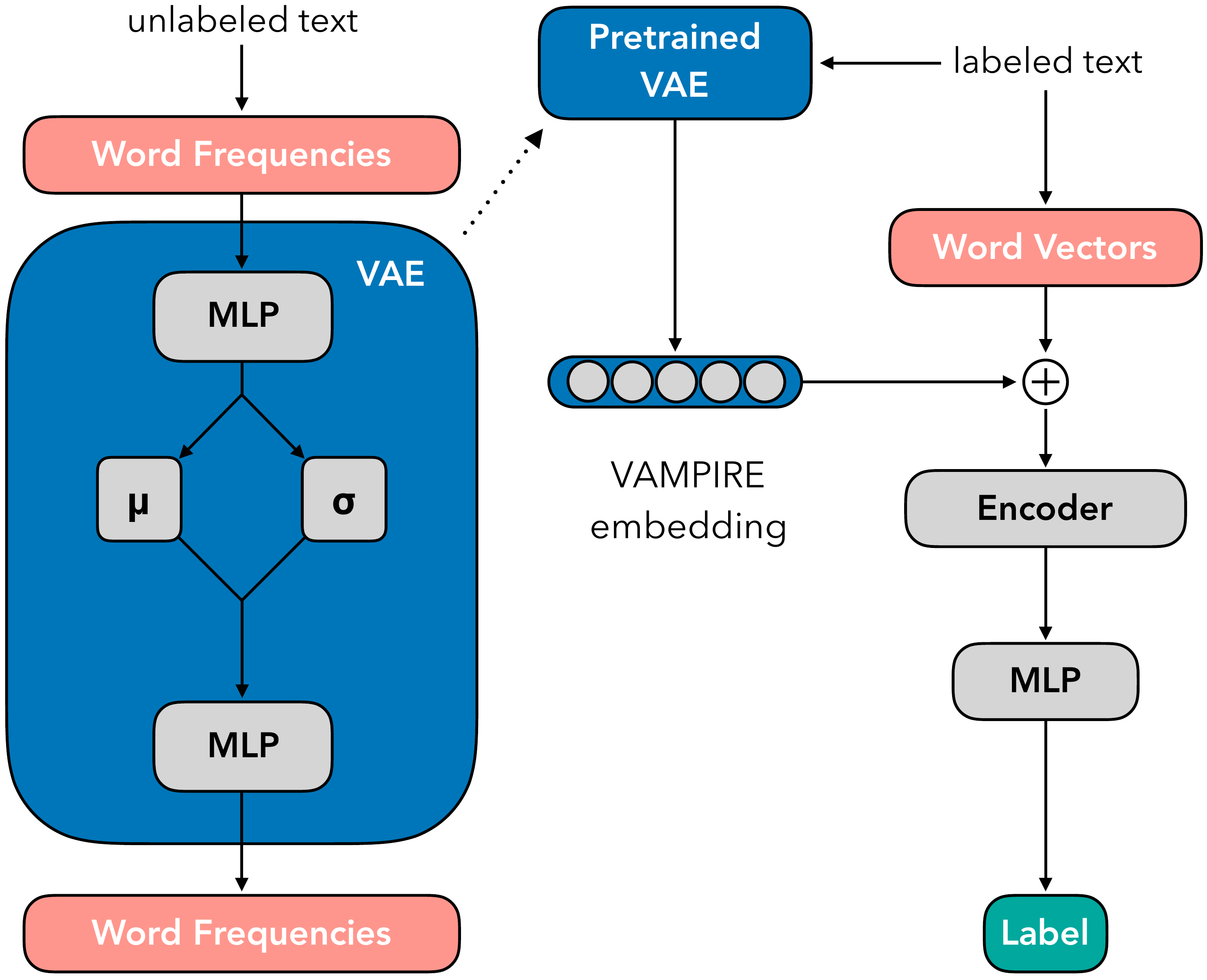}
    \caption{\VAMPIRE~involves pretraining a deep variational autoencoder (VAE; displayed on left) on unlabeled text. The VAE, which consists entirely of feedforward networks, learns to reconstruct a word frequency representation of the unlabeled text with a logistic normal prior, parameterized by $\bm{\mu}$ and $\bm{\sigma}$. Downstream, the pretrained VAE's internal states are frozen and concatenated to task-specific word vectors to improve classification in the low-resource setting.}
    \label{fig:schematic}
\end{figure}

Our approach heavily borrows from past work on VAEs \citep{kingma.2013,miao.2016,Srivastava2017AutoencodingVI}, which we adapt to semi-supervised text classification (see Figure \ref{fig:schematic}). We do so by pretraining the document model on unlabeled data (\S\ref{sec:pretraining}), and then using learned representations in a downstream classifier (\S\ref{sec:classification}). The downstream classifier makes use of multiple internal states of the pretrained document model, as in \citet{Peters2018DissectingCW}. We also explore how to best do model selection in a way that benefits the downstream task (\S\ref{sec:modelselection}).

\subsection{Unsupervised Pretraining} \label{sec:pretraining}

In order to learn useful representations, we initially ignore labels, and assume each document is generated from a latent variable, $\bm{z}$. The functions learned in estimating this model then provide  representations which are  used as features in supervised learning. 

Using a variational autoencoder for approximate Bayesian inference, we simultaneously learn an \textit{encoder}, which maps from the observed text to an approximate posterior $q(\bm{z} \mid \bm{x})$, and a \textit{decoder}, which reconstructs the text from the latent representation. In practice, we instantiate both the encoder and decoder as neural networks and assume that the encoder maps to a normally distributed posterior, 
i.e., for document $i$,
\begin{align}
q(\bm{z}_i \mid \bm{x}_i) &= \mathcal{N}\left(\bm{z}_i \mid f_\mu(\bm{x}_i),  \textrm{diag}(f_\sigma(\bm{x}_i)) \right) \\
\bm{x}_i &\sim p(\bm{x}_i \mid f_d(\bm{z}_i)).
\end{align}

Using standard principles of variational inference, we derive a variational bound on the marginal log-likelihood of the observed data,
\begin{equation}
\begin{split}
\log p(\bm{x}_i) \geq \mathcal{B}(\bm{x}_i) = \mathbb{E}_{q(\bm{z}_i \mid \bm{x}_i)} 
   [ \log p(\bm{x}_{i} \mid \bm{z}_i ) ] \\
   -\textrm{KL}[q(\bm{z}_i \mid \bm{x}_i) ~\|~ p(\bm{z}) ].
\end{split}
\label{eq:bound}
\end{equation}

Intuitively, the first term in the bound can be thought of as a \emph{reconstruction loss}, ensuring that generated words are similar to the original document. The second term,  the \emph{KL divergence}, encourages the variational approximation to be close to the assumed prior, $p(\bm{z})$, which we take to be a spherical normal distribution.

Using the \textit{reparameterization trick} \citep{kingma.2013,rezende.2014}, we replace the expectation with a single-sample approximation,\footnote{We leave experimentation with multi-sample approximation (e.g., importance sampling) to future work.} i.e.,
\begin{align}
    \mathcal{B}(\bm{x}_i) &\approx \log p(\bm{x}_{i} \mid \bm{z}_i^{(s)})
   -\textrm{KL}[q(\bm{z}_i \mid \bm{x}_i) ~\|~ p(\bm{z}) ] \\
  \bm{z}_i^{(s)} &= f_\mu(\bm{x}_i) + f_\sigma(\bm{x}_i) \cdot \bm{\varepsilon}^{(s)},
\end{align}
where $\bm{\varepsilon}^{(s)} \sim \mathcal{N} (0, \mathbf{I})$ is sampled from an independent normal. 
All parameters can then be optimized simultaneously by performing stochastic gradient ascent on the variational bound. 

A powerful way of encoding and decoding text is to use sequence models. That is, $f_\mu(\bm{x})$ and $f_\sigma(\bm{x})$ would map from a sequence of tokens to a pair of vectors, $\bm{\mu}$ and $\bm{\sigma}$, and $f_d(\bm{z})$ would similarly decode from $\bm{z}$ to a sequence of tokens, using recurrent, convolutional, or attention-based networks. Some authors have adopted this approach \cite{bowman.2016,xu.2017,yang.2017}, but as discussed above (\S \ref{sec:ssmodels}), it has a number of disadvantages.

In this paper, we adopt a more lightweight and directly interpretable approach, and work with word frequencies instead of word sequences. Using the same basic structure as \citet{miao.2016} but employing a softmax in the decoder, we encode $f_\mu(\boldsymbol{x})$ and $f_\sigma(\boldsymbol{x})$ with multi-layer feed forward neural networks operating on an input vector of word counts, $\bm{c}_i$:
\begin{align}
\bm{c}_i &= \textrm{counts}(\bm{x}_i)  \label{eq:counts} \\
\bm{h}_i &= \textrm{MLP}(\bm{c}_i) \\
\bm{\mu}_i &= f_\mu(\bm{x}_i) = \bm{W}_\mu \bm{h}_i + \bm{b}_\mu \\
\bm{\sigma}_i &= f_\sigma(\bm{x_i}) = \exp(\bm{W}_\sigma \bm{h}_i + \bm{b}_\sigma ) \\
\bm{z}_i^{(s)} &= \bm{\mu}_i + \bm{\sigma}_i \cdot \bm{\varepsilon}^{(s)}.
\end{align}

For a decoder, we use the following form, which reconstructs the input in terms of topics (coherent distributions over the vocabulary):
\begin{align}
    \bm{\theta}_i &= \textrm{softmax}(\bm{z}_i^{(s)}) \\
    \bm{\eta}_i &= \textrm{softmax}(\bm{b} + \bm{B}\bm{\theta}_i) \\
    \log p(\bm{x}_i \mid \bm{z}_i^{(s)}) &= \sum_{j=1}^{V} \bm{c}_{ij} \cdot \log \bm{\eta}_{ij},
\end{align}
where $j$ ranges over the vocabulary.

By placing a softmax on $\bm{z}$, we can interpret $\bm{\theta}$ as a distribution over latent topics, as in a topic model \citep{blei.2003}, and $\bm{B}$ as representing positive and negative topical deviations from a background $\bm{b}$. This form (essentially a unigram LM) allows for much more efficient inference on $\bm{z}$, compared to sequence-based encoders and decoders.

\subsection{Model Selection via Topic Coherence}
\label{sec:modelselection}

Because our pretraining ignores document labels, it is not obvious that optimizing it to convergence will produce the best representations for downstream classification.
When pretraining using a LM objective, models are typically trained until model fit stops improving (i.e., perplexity on validation data).
In our case, however, $\bm{\theta}_i$ has a natural interpretation as the distribution (for document $i$) over the latent ``topics'' learned by the model ($\bm{B}$). As such, an alternative is to use the quality of the topics as a criterion for early stopping.

It has repeatedly been observed that different types of topic models offer a trade-off between perplexity and topic quality  \citep{chang.2009,Srivastava2017AutoencodingVI}.
Several methods for automatically evaluating topic \textit{coherence} have been proposed \citep{newman.2010,mimno.2011}, such as normalized pointwise mutual information (NPMI), which \citet{lau.2014} found to be among the most strongly correlated with human judgement.  As such, we consider using either log likelihood or NPMI as a stopping criteria for \VAMPIRE pretraining (\S\ref{sec:stopping}), and evaluate them in terms of which leads to the better downstream classifier.

NPMI measures the probability that two words collocate in an external corpus (in our case, the validation data). For each topic $t$ in $\bm{B}$, we collect the top ten most probable words and compute NPMI between all pairs:
\begin{align}
 \textrm{NPMI}(t) &= \sum_{\substack{i,j \le 10;~j \neq i}} \frac{\log \frac{P(t_i, t_j)} {P(t_i)P(t_j)}} {-\log P(t_i, t_j)}
\end{align}
We then arrive at a global NPMI for $\bm{B}$ by averaging the NPMIs across all topics. We evaluate NPMI at the end of each epoch during pretraining, and stop training when NPMI has stopped increasing for a pre-defined number of epochs. 

\subsection{Using a Pretrained VAE for Text Classification} \label{sec:classification}

\citet{kingma.2014} proposed using the latent variable of an unsupervised VAE as features in a downstream model for classifying images.
However, work on pretraining for NLP, such as \citet{peters.2018}, found that LMs encode different information in different layers, each of which may be more or less useful for certain tasks. Here, for an $n$-layer \textbf{MLP} encoder on word counts $\bm{c}_i$, we build on that idea, and use as representations a weighted sum over $\bm{\theta}_i$ and the internal states of the MLP, $\bm{h}_{i}^{(k)}$, with weights to be learned by the downstream classifier.\footnote{We also experimented with the joint training and combined approaches discussed in \citet{kingma.2014}, but found that neither of these reliably improved performance over our pretraining approach.}

That is, for any sequence-to-vector encoder, $f_{\textit{s2v}}(\bm{x})$, we propose to augment the vector representations for each document by concatenating them with a weighted combination of the internal states of our variational encoder \cite{peters.2018}. We can then train a supervised classifier on the weighted combination,
\begin{align}
    \bm{r}_i &= \lambda_0 \bm{\theta}_i + \sum_{k=1}^n \lambda_k \bm{h}_i^{(k)}  \\
    p(y_i \mid \bm{x}_i) &= f_c( [\bm{r}_i; f_{s2v}(\bm{x_i})]),
\end{align}
where $f_c$ is a neural classifier and  $\{\lambda_0, \ldots, \lambda_n\}$ are softmax-normalized trainable parameters.

\subsection{Optimization} \label{sec:optimization} 

In all cases, we optimize models using Adam \citep{adam}. In order to prevent divergence during pretraining, we make use of a batch-norm layer on the reconstruction of $\bm{x}$ \cite{ioffe.2015}. We also use KL-annealing \citep{bowman.2016}, placing a scalar weight on the KL divergence term in Eq.(\ref{eq:bound}), which we gradually increase from zero to one.
Because our model consists entirely of feedforward neural networks, it is easily parallelized, and can run efficiently on either CPUs or GPUs.

\section{Experimental Setup}

We evaluate the performance of our approach on four text classification tasks, as we vary the amount of labeled data, from 200 to 10,000 instances.
In all cases, we assume the existence of about 75,000 to 125,000 unlabeled in-domain examples, which come from the union of the unused training data and any additional unlabeled data provided by the corpus.
Because we are working with a small amount of labeled data, we run each experiment with five random seeds, each with a different sample of labeled training instances, and report the mean performance on test data.

\subsection{Datasets and Preprocessing}

We experiment with text classification datasets that span a variety of label types. The datasets we use are the familiar \ag~News \citep{zhang.2015}, \imdb~\citep{maas.2011}, and \yahoo~Answers datasets \cite{chang.2008}, as well as a dataset of tweets labeled in terms of four \hatespeech~categories \citep{founta.2018}. Summary statistics are presented in Table \ref{tab:table1}.
In all cases, we either use the official test set, or take a random stratified sample of 25,000 documents as a test set. We also sample 5,000 instances as a validation set.

We tokenize documents with spaCy, and use up to 400 tokens for sequence encoding ($f_{s2v}(\bm{x})$). For \VAMPIRE~pretraining, we restrict the vocabulary to the 30,000 most common words in the dataset, after excluding tokens shorter than three characters, those with digits or punctuation, and stopwords.\footnote{\url{http://snowball.tartarus.org/algorithms/english/stop.txt}} We leave the vocabulary for downstream classification unrestricted.

\begin{table}[t]
    \centering
    \small
    \begin{tabular}{llcc}
    \toprule
    \textbf{Dataset}   &  \textbf{Label Type}  & \textbf{Classes} & \textbf{Documents}  \\
    \midrule
    \ag & topic &  4 &  127600 \\ 
    \hatespeech & hatespeech  & 4 & 99996  \\   
    \imdb & sentiment &  2 & 100000 \\
    \yahoo & topic & 15 & 150015 \\    
    \bottomrule
    \end{tabular}
    \caption{Datasets used in our experiments.}
    \label{tab:table1}
\end{table}

\subsection{\VAMPIRE~Architecture}

In order to find reasonable hyperparameters for \VAMPIRE, we utilize a random search strategy for pretraining. For each dataset, we take the model with the best NPMI for use in the downstream classifiers.  We detail sampling bounds and final assignments for each hyperparameter in Table \ref{tab:vampire_hyperparameters_imdb} in Appendix \ref{sec:vampiresearch}.

\subsection{Downstream Classifiers}

For all experiments we make use of the Deep Averaging Network (DAN) architecture \citep{iyyer.2015} as our baseline sequence-to-vector encoder, $f_{s2v}(\bm{x})$. That is, embeddings corresponding to each token are summed and passed through a multi-layer perceptron.
\begin{equation}
\begin{split}
    p(y_i \mid \bm{x}_i) = \textrm{MLP} \left( \frac{1}{|\bm{x}_i|} \textstyle \sum_{j=1}^{|\bm{x}_i|}E(\bm{x}_{i})_j \right),
\end{split}
\label{eq:DAN}
\end{equation}
where $E(\bm{x})$ converts a sequence of tokens to a sequence of vectors, using randomly initialized vectors, off-the-shelf \glove~embeddings \citep{glove}, or contextual embeddings.

To incorporate the document representations learned by \VAMPIRE in a downstream classifier, we concatenate them with the average of randomly initialized trainable embeddings, i.e., 
\begin{align}
    p(y_i \mid \bm{x}_i) = \textrm{MLP} \left( \left[ \bm{r}_i ; \frac{1}{|\bm{x}_i|} \textstyle \sum_{j=1}^{|\bm{x}_i|}E(\bm{x}_{i})_j \right] \right).
\end{align}

Preliminary experiments found that DANs with one-layer MLPs and moderate dropout provide more reliable performance on validation data than more expressive models, such as CNNs or LSTMs, with less hyperparameter tuning, especially when working with few labeled instances (details in Appendix \ref{sec:classifiersearch}).

\subsection{Resources and Baselines} \label{sec:resources}

In these experiments, we consider baselines for both  low-resource and high-resource settings, where the high-resource baselines have access to greater computational resources and a either massive amount of unlabeled data or a pretrained model, such as \elmo~or \bert.\footnote{As discussed above, we consider these models to be representative of the high-resource setting, both because they were computationally intensive to train, and because they were made possible by the huge amount of English text that is available online.}

\paragraph{Low resource}   In the low-resource setting we assume that computational resources are at a premium, so we are limited to lightweight approaches such as \VAMPIRE, which can run efficiently on a CPU. 
As baselines, we consider \textbf{a)} a purely supervised model, with randomly initialized 50-dimensional embeddings and no access to unlabeled data; \textbf{b)} the same model initialized with 300-dimensional \glove~vectors, pretrained on 840 billion words;\footnote{\url{http://nlp.stanford.edu/projects/glove/}} \textbf{c)} 300-dimensional \glove~vectors trained on only in-domain data; and \textbf{d)} self-training, which has access to the in-domain unlabeled data. For self-training, we iterate over training a model, predicting labels on all unlabeled instances, and adding to the training set all unlabeled instances whose label is predicted with high confidence, repeating this up to five times and using the model with highest validation accuracy.
On each iteration, the threshold for a given label is equal to the 90th percentile of predicted probabilities for validation instances with the corresponding label.

\begin{figure}[t]
    \centering
    \includegraphics[scale=0.48]{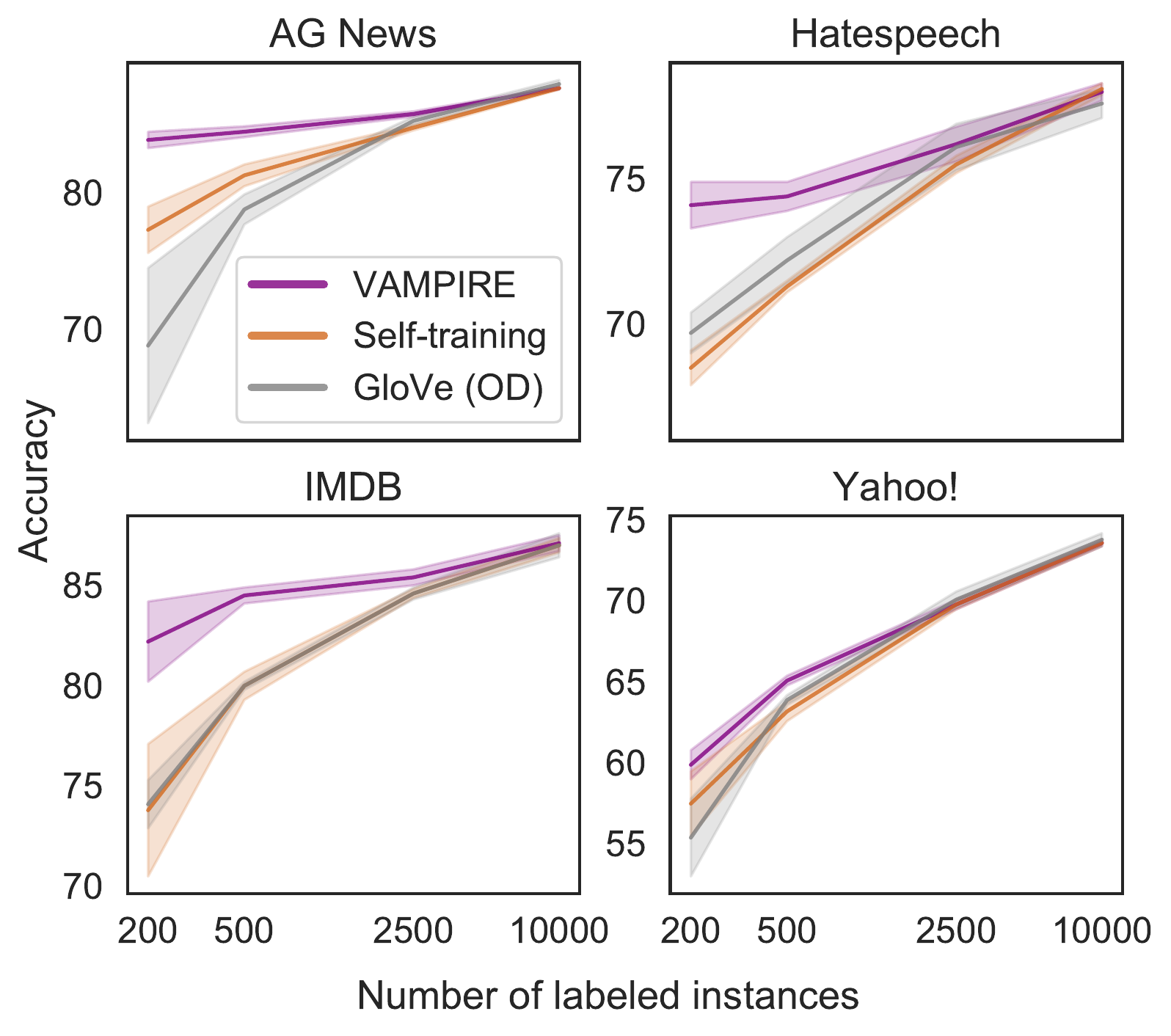}
    \caption{Learning curves for all datasets in the low-resource setting, showing the mean (line) and one standard deviation (bands) over five runs for \VAMPIRE, self-training, and 840B-token \glove~embeddings. Full results are in Table \ref{tab:results}.}
    \label{fig:bands}
\end{figure}

\paragraph{High resource}

In the \textit{high-resource} setting, we assume access to plentiful computational resources and massive amounts of out-of-domain data, which may be indirectly accessed through pretrained models.  Specifically, we evaluate the performance of a Transformer-based \elmo~\cite{Peters2018DissectingCW} and \bert, both (a) off-the-shelf with frozen embeddings and (b) after semi-supervised fine-tuning to both unlabeled and  labeled in-domain data. To perform semi-supervised fine-tuning, we first use \elmo~and \bert's original objectives to fine-tune to the unlabeled data. To fine-tune \elmo~to the labeled data, we average over the LM states and add a softmax classification layer. We obtain the best results applying slanted triangular learning rates and gradual unfreezing \cite{ruder.2018} to this fine-tuning step. To fine-tune \bert~to labeled data, we feed the hidden state corresponding to the \texttt{[CLS]} token of each instance to a softmax classification layer. We use AllenNLP\footnote{\url{https://allennlp.org/elmo}} to fine-tune \elmo, and Pytorch-pretrained-BERT\footnote{\url{https://github.com/huggingface/pytorch-pretrained-BERT}} to fine-tune \bert.

We also experiment with \elmo~trained only on in-domain data as an example of high-resource LM pretraining methods, such as \citet{dai.2015}, when there is no out-of-domain data available. Specifically, we generate contextual word representations with a Transformer-based \elmo.
During downstream classification, the resulting vectors are frozen and concatenated to randomly initialized word vectors prior to the summation in Eq. (\ref{eq:DAN}).

\section{Results}
\label{sec:results}

In the \textbf{low-resource} setting, we find that \VAMPIRE~achieves the highest accuracy of all low-resource methods we consider, especially when the amount of labeled data is small.
Table \ref{tab:results} shows the performance of all low-resource models on all datasets as we vary the amount of labeled data, and a subset of these are also shown in Figure \ref{fig:bands} for easy comparison.

\begin{table*}[t!]
\small
\centering
\begin{tabular}{lrrrrr}
  \toprule
  \textbf{Dataset} & \textbf{Model} & \textbf{200} &  \textbf{500} &  \textbf{2500}  & \textbf{10000} \\
  \midrule
  \imdb  & Baseline  & 68.5 (7.8) & 79.0 (0.4) & 84.4 (0.1) & 87.1 (0.3) \\
      & Self-training & 73.8 (3.3) & 80.0 (0.7) & 84.6 (0.2) & 87.0 (0.4) \\
      & \glove~(ID) & 74.5 (0.8) & 79.5 (0.4) & 84.7 (0.2) & 87.1 (0.4) \\
      & \glove~(OD) & 74.1 (1.2) & 80.0 (0.2) & 84.6 (0.3) & 87.0 (0.6) \\ 
      & \VAMPIRE  & \textbf{82.2} (2.0) & \textbf{84.5} (0.4) & \textbf{85.4} (0.4) & \textbf{87.1} (0.4) \\
    
  \midrule
  \midrule
  \ag & Baseline  & 68.8 (2.0) & 77.3 (1.0) & 84.4 (0.1) & 87.5 (0.2) \\
      & Self-training & 77.3 (1.7) & 81.3 (0.8) & 84.8 (0.2) & 87.7 (0.1) \\
      & \glove~(ID) & 70.4 (1.2) & 78.0 (1.0) & 84.1 (0.3) & 87.1 (0.2) \\
      & \glove~(OD) & 68.8 (5.7) & 78.8 (1.1) & 85.3 (0.3) & \textbf{88.0} (0.3) \\     
      & \VAMPIRE  & 
      \textbf{83.9} (0.6) & \textbf{84.5} (0.4) & \textbf{85.8} (0.2) & 87.7 (0.1)\\
      
  \midrule
  \midrule
  \yahoo & Baseline  & 54.5 (2.8) & 63.0 (0.5) & 69.5 (0.3) & 73.6 (0.2) \\
      & Self-training & 57.5 (2.0) & 63.2 (0.6) & 69.8 (0.3) & 73.6 (0.2) \\
      & \glove~(ID) & 55.2 (2.3) & 63.5 (0.3) & 69.7 (0.3) & 73.5 (0.3) \\ 
      & \glove~(OD)  & 55.4 (2.4) & 63.9 (0.3) & \textbf{70.1} (0.5) & \textbf{73.8} (0.4)\\      
      & \VAMPIRE  & 
      \textbf{59.9} (0.9) & \textbf{65.1} (0.3) & 69.8 (0.3) & 73.6 (0.2)\\

  \midrule
  \midrule
  \hatespeech &  Baseline  & 67.7 (1.8) & 71.3 (0.2) & 75.6 (0.4) & 77.8 (0.2) \\
             & Self-training & 68.5 (0.6) & 71.3 (0.2) & 75.5 (0.3) & 78.1 (0.2) \\
        & \glove~(ID) & 69.7 (1.2) & 71.9 (0.5) & 76.0 (0.3) & \textbf{78.3} (0.2) \\
        & \glove~(OD) & 69.7 (0.7) & 72.2 (0.8) & 76.1 (0.8) & 77.6 (0.5) \\
        & \VAMPIRE  & \textbf{74.1} (0.8) & \textbf{74.4} (0.5) & \textbf{76.2} (0.6) & 78.0 (0.3)\\

  \bottomrule
\end{tabular}
\caption{Test accuracies in the low-resource setting on four text classification datasets under varying levels of labeled training data (200, 500, 2500, and 10000 documents). Each score is reported as an average over five seeds, with standard deviation in parentheses, and the highest mean result in each setting shown in bold.}
\label{tab:results}
\end{table*}

\begin{figure}[t]
    \centering
    \includegraphics[scale=0.47]{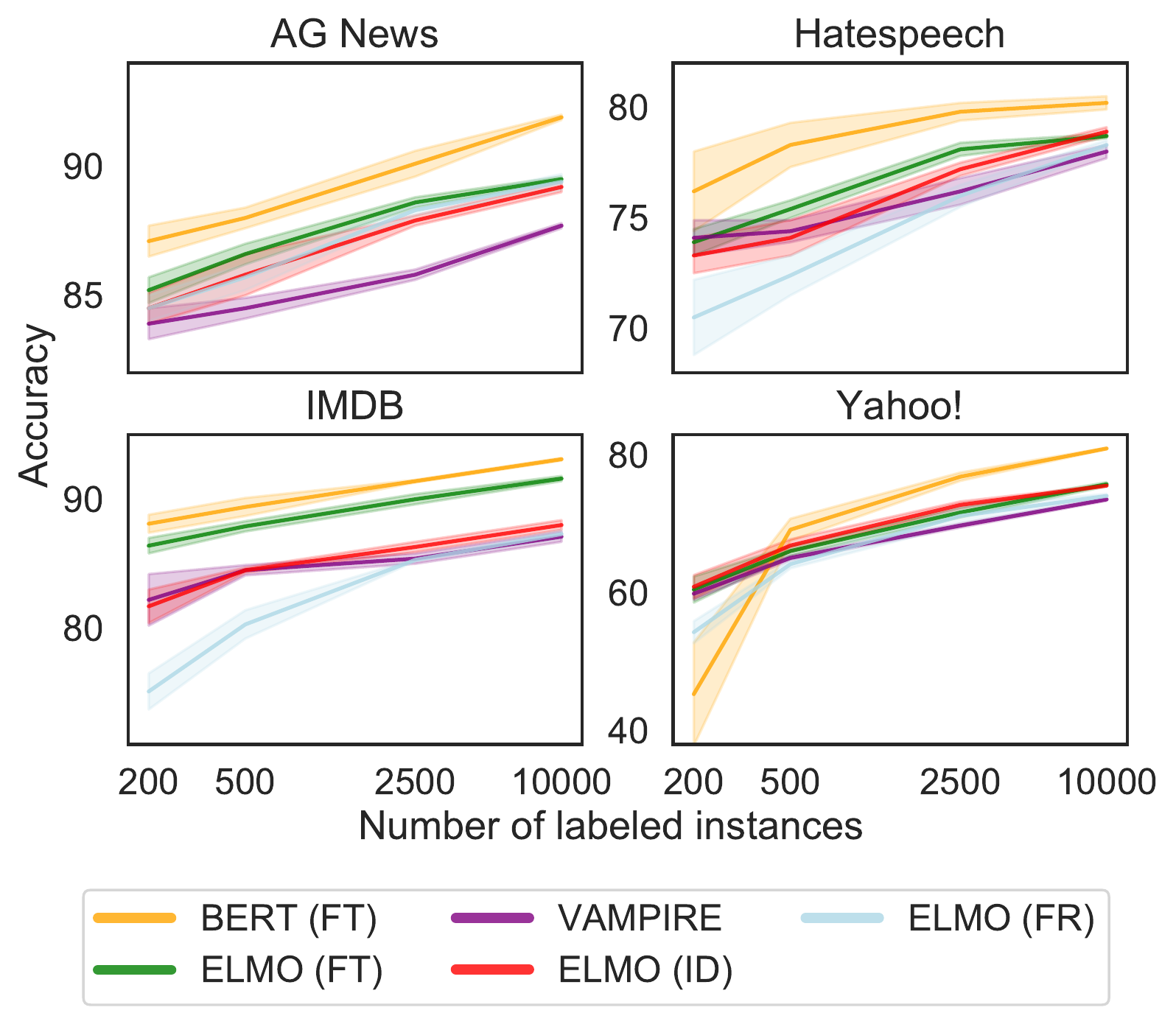}
    \caption{High-resource methods (plus \VAMPIRE) on four datasets; \elmo~performance benefits greatly from training on (ID), or fine-tuning (FT) to, the in-domain data (as does \bert; full results in Appendix \ref{sec:appendix_results}). Key: FT (fine-tuned), FR (frozen), ID (in-domain).
    }
    \label{fig:high-resource}
\end{figure}

In the \textbf{high-resource} setting, we find, not surprisingly, that fine-tuning the pretrained \bert~model to in-domain data provides the best performance. For both \bert~and \elmo, we find that using frozen off-the-shelf vectors results in surprisingly poor performance, compared to fine-tuning to the task domain, especially for \hatespeech~and \imdb.\footnote{See also \citet{ruder.2018}.} For these two datasets, an \elmo~model trained \textit{only} on in-domain data offers far superior performance to frozen off-the-shelf \elmo~(see Figure \ref{fig:high-resource}).
This difference is smaller, however, for \yahoo~and \ag. (Please see Appendix \ref{sec:appendix_results} for full results).

These results taken together demonstrate that although pretraining on massive amounts of web text offers large improvements over purely supervised models, access to unlabeled \textit{in-domain} data is critical, either for fine-tuning a pretrained language model in the high-resource setting, or for training \VAMPIRE in the low-resource setting. Similar findings have been reported by \citet{Yogatama2019LearningAE} for tasks such as natural language inference and question answering.

\begin{figure}[t!]
    \hspace*{-3.5mm}
    \includegraphics[scale=0.50]{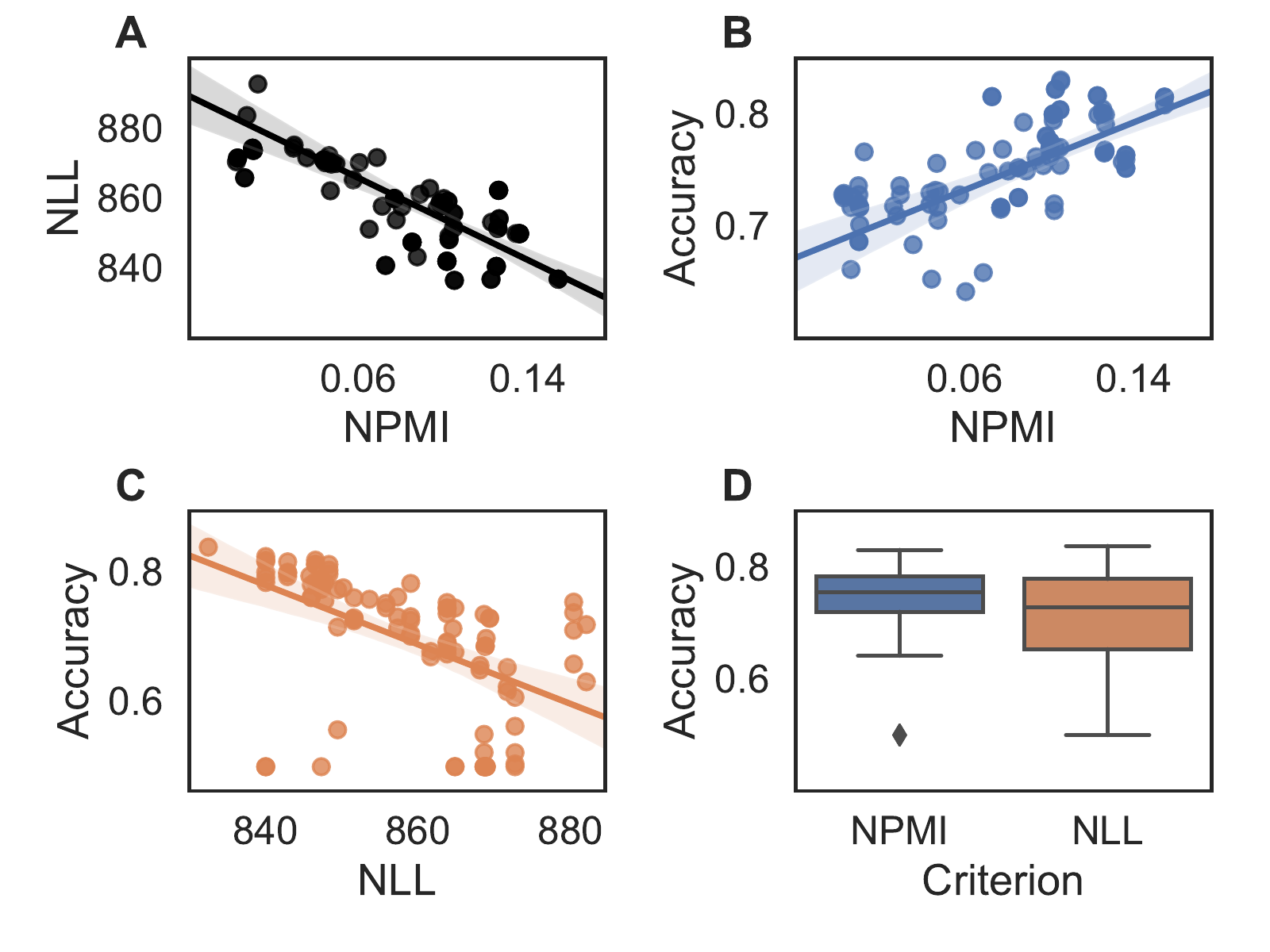}
    \caption{Comparing NPMI and NLL as early stopping criteria for \VAMPIRE model selection. NPMI and NLL are correlated measures of model fit, but NPMI-selected \VAMPIRE models have lower variance on downtream classification performance with 200 labeled documents of \imdb. Accuracy is reported on the validation data. See \S\ref{sec:stopping} for more details.}
    \label{fig:npmi_nll}
\end{figure}

\section{Analysis}

\subsection{NPMI versus NLL as Stopping Criteria} \label{sec:stopping}

To analyze the effectiveness of different stopping criterion in \VAMPIRE, we pretrain 200 \VAMPIRE models on \imdb: 100 selected via NPMI, and 100 selected via negative log likelihood (NLL) on validation data. Interestingly, we observe that \VAMPIRE NPMI and NLL values are negatively correlated ($\rho$ = --0.72; Figure~\ref{fig:npmi_nll}A), suggesting that upon convergence, trained models that better fit the data also tend to have more coherent topics. We then train 200 downstream classifiers with the same hyperparameters, on a fixed 200 document random subset of the IMDB dataset, uniformly sampling over the NPMI- and NLL-selected \VAMPIRE models as additional features. In Figure~\ref{fig:npmi_nll}B and Figure~\ref{fig:npmi_nll}C, we observe that better pretrained \VAMPIRE models (according to either criterion) tend to produce better downstream performance.
($\rho$ = 0.55 and $\rho$ = --0.53, for NPMI and NLL respectively).

However, we also observe higher variance in accuracy among the \VAMPIRE models obtained using NLL as a stopping criterion
(Figure~\ref{fig:npmi_nll}D). Such models selected via NLL have poor topic coherence and downstream performance. As such, doing model selection using NPMI is the preferred alternative, and all \VAMPIRE results in Table \ref{tab:results} are based on pretrained models selected using this criterion.

The experiments in \citet{ding.2018} provide some insight into this behaviour. They find that when training neural topic models, model fit and NPMI initially tend to improve on each epoch. At some point, however, perplexity continues to improve, while NPMI starts to drop, sometimes dramatically. We also observe this phenomenon when training \VAMPIRE (see Appendix \ref{sec:appendix_stopping}). Using NPMI as a stopping criterion, as we propose to do, helps to avoid degenerate models that result from training too long.

In some preliminary experiments, we also observe cases where NPMI is artificially high because of redundancy in topics. Applying batchnorm to the  reconstruction markedly improves diversity of collocating words across topics, which has also been noted by \citet{Srivastava2017AutoencodingVI}. Future work may explore assigning a word diversity regularizer to the NPMI metric, so as to encourage models that have both stronger coherence and word diversity across topics.

\subsection{Learned Latent Topics}

In addition to being lightweight, one advantage of \VAMPIRE is that it produces document representations that can be explicitly interpreted in terms of topics. Although the input we feed into the downstream classifier combines this representation with internal states of the encoder, the topical interpretation helps to summarize what the pretraining has learned. Examples of topics learned by \VAMPIRE are provided in Table~\ref{tab:topics} and Appendix \ref{sec:additionaltopics}.

\begin{table}[t!]
    \small
    \centering
    \begin{tabular}{cccc}

        \multicolumn{2}{c}{\imdb} & \multicolumn{2}{c}{\yahoo}\\
        
        \toprule

        \textbf{Horror} & \textbf{Classics} & \textbf{Food} & \textbf{Obstetrics}\\
        
        \midrule
        giallo & dunne & cuisine & obstetrics \\
         horror & cary & peruvian & vitro \\ 
         gore & abbott & bengali & endometriosis \\
         lugosi & musicals & cajun & fertility \\
         zombie & astaire & potato & contraceptive \\
         dracula & broadway & carne & pregnancy \\
         bela & irene & idli & birth \\
         cannibal & costello & pancake & ovarian \\
         vampire & sinatra & tofu & menstrual \\
         lucio & stooges & gumbo & prenatal \\
        \bottomrule

    \end{tabular}
    \caption{Example topics learned by \VAMPIRE~in \imdb~and \yahoo~datasets. See Appendix \ref{sec:additionaltopics} for more examples.}
    \label{tab:topics}
\end{table}

\subsection{Learned Scalar Layer Weights}

Since the scalar weight parameters in $\bm{r}_i$ are trainable, we are able to investigate which layers of the pretrained VAE the classifier tends to prefer. We consistently find that the model tends to upweight the first layer of the VAE encoder, $\bm{h}^{(1)}$, and $\bm{\theta}$, and downweight the other layers of the encoder. To improve learning, especially under low resource settings, we initialize the scalar weights applied to the first encoder layer and $\bm{\theta}$ with high values and downweighted the intermediate layers, which increases validation performance. However, we also have  observed that using a multi-layer encoder in \VAMPIRE leads to larger gains downstream.

\subsection{Computational Requirements}
An appealing aspect of \VAMPIRE~is its compactness. Table \ref{tab:speed} shows the computational requirements involved in training \VAMPIRE on a single GPU or CPU, compared to training an \elmo~model from scratch on the same data on a GPU. It is possible to train \VAMPIRE~orders of magnitude faster than \elmo, even without expensive hardware, making it especially suitable for obtaining fast results when resources are limited.

\begin{table}[t!]
    \small
    \centering
    \hspace*{-2.5mm}
    \begin{tabular}{lrr}
        \toprule

        \textbf{Model} & \textbf{Parameters} & \textbf{Time} \\
        \midrule
        \VAMPIRE (GPU) & 3.8M & 7 min\\
        \VAMPIRE (CPU)  & 3.8M & 22 min\\
         \elmo~(GPU) & 159.2M & 12 hr 35 min \\
         \bottomrule

    \end{tabular}
\caption{\VAMPIRE~is substantially more compact than Transformer-based \elmo~but is still competitive under low-resource settings. Here, we display the computational requirements for pretraining \VAMPIRE and \elmo~on in-domain unlabeled text from the IMDB dataset. We report results on training \VAMPIRE~(with hyperparameters listed in Appendix \ref{sec:vampiresearch}) and \elmo~(with its default configuration) on a GeForce GTX 1080 Ti GPU, and \VAMPIRE on a 2.60GHz Intel Xeon CPU. VAMPIRE uses about 750MB of memory on a GPU, while \elmo~requires about 8.5GB.}
    \label{tab:speed}
\end{table}

\section{Related Work}

In addition to references given throughout, many others have explored ways of enhancing performance when working with limited amounts of labeled data. Early work on speech recognition demonstrated the importance of pretraining and fine-tuning deep models in the semi-supervised setting \citep{dong.2010}. \citet{chang.2008}  considered ``dataless'' classification, where the names of the categories provide the only supervision. \citet{miyato.2016} showed that adversarial pretraining can offer large gains, effectively augmenting the amount of data available.  A long line of work in \textit{active learning} similarly tries to maximize performance when obtaining labels is costly \citep{settles.2010}. \citet{Xie2019UnsupervisedDA} describe novel data augmentation techniques leveraging back translation and tf-idf word replacement. All of these approaches could be productively combined with the methods proposed in this paper.

\section{Recommendations} \label{sec:discussion}

Based on our findings in this paper, we offer the following practical advice to those who wish to do effective semi-supervised text classification.
\begin{itemize}
    \item When resources are unlimited, the best results can currently be obtained by using a pretrained model such as \bert, but fine-tuning to in-domain data is critically important \citep[see also][]{ruder.2018}.
    \item When computational resources and annotations are limited, but there is plentiful unlabeled data, \VAMPIRE offers large gains over other low-resource approaches.
    \item Training a language model such as \elmo~on only in-domain data offers comparable or somewhat better performance to \VAMPIRE, but may be prohibitively expensive, unless working with GPUs.
    \item Alternatively, resources can be invested in getting more annotations; with sufficient labeled data  (tens of thousands of instances), the advantages offered by additional unlabeled data become negligible. Of course, other NLP tasks may involve different trade-offs between data, speed, and accuracy.
\end{itemize}

\section{Conclusions}

The emergence of models like \elmo~and \bert~has revived semi-supervised NLP, demonstrating that pretraining large models on massive amounts of data can provide representations that are beneficial for a wide range of NLP tasks. In this paper, we confirm that these models are useful for text classification when the number of labeled instances is small, but demonstrate that fine-tuning to in-domain data is also of critical importance. 
In settings where \bert~cannot easily be used, either due to computational limitations, or because an appropriate pretrained model in the relevant language does not exist, \VAMPIRE offers a competitive lightweight alternative for pretraining from unlabeled data in the low-resource setting.
When working with limited amounts of labeled data, we achieve superior performance to baselines such as self-training, or using word vectors pretrained on out-of-domain data, 
and approach the performance of \elmo~trained only on in-domain data at a fraction of the computational cost.

\section*{Acknowledgments}

We thank the members of the AllenNLP and ARK teams for useful comments and discussions. We also thank the anonymous reviewers for their insightful feedback. Computations on \url{beaker.org} were supported in part by credits from Google Cloud.

\bibliography{acl2019}
\bibliographystyle{acl_natbib}

\clearpage

\appendix

\section{Hyperparameter Search}
\label{sec:hyperparam_appendix}

In this section, we describe the hyperparameter search we used to choose model configurations, and include plots illustrating the range of validation performance observed in each setting.

\begin{table*}[t!]
    \centering
    \small

    \begin{tabular}{cc}
      \toprule
      \textbf{Computing Infrastructure} & GeForce GTX 1080 GPU\\ 
      \midrule
      \textbf{Number of search trials} & 60 trials per dataset \\
      \midrule
      \textbf{Search strategy} & uniform sampling \\ 
      \midrule
      \textbf{Model implementation} & \url{http://github.com/allenai/vampire}\\
      \bottomrule
    \end{tabular}

    \vspace{3mm}\begin{tabular}{llllll}
        
        \toprule
        \textbf{Hyperparameter} & \textbf{Search space} & \textbf{\imdb} &  \textbf{\ag}  & \textbf{\yahoo}   & \textbf{\hatespeech}  \\
        \midrule
        number of epochs & 50  & 50 & 50 & 50 & 50 \\
        \midrule
        patience & 5  & 5 & 5 & 5 & 5\\
        \midrule
        batch size & 64  &  64 & 64 & 64 & 64\\
        \midrule
        KL divergence annealing & \emph{choice}[sigmoid, linear, constant] & linear & linear & linear & constant\\
        \midrule
        KL annealing sigmoid weight 1 &  0.25 & N/A & N/A & N/A & N/A \\
        \midrule 
        KL annealing sigmoid weight 2 & 15 & N/A & N/A & N/A & N/A \\
        \midrule
        KL annealing linear scaling & 1000 &  1000 & 1000 & 1000 & N/A  \\
        \midrule
        VAMPIRE hidden dimension & \emph{uniform-integer}[32, 128] & 80 & 81 & 118 & 125 \\
        \midrule
        Number of encoder layers & \emph{choice}[1, 2, 3] & 2 & 2 & 3 & 3 \\
        \midrule
        Encoder activation & \emph{choice}[relu, tanh, softplus] & tanh & relu & tanh & softplus \\
        \midrule
        Mean projection layers & 1 & 1 & 1 & 1 & 1\\
        \midrule
        Mean projection activation & linear & linear & linear & linear & linear \\
        \midrule
        Log variance projection layers & 1 & 1 & 1 & 1 & 1 \\
        \midrule
        Log variance projection activation & linear & linear & linear & linear & linear \\
        \midrule
        Number of decoder layers & 1 & 1 & 1 & 1 & 1 \\
        \midrule
        Decoder activation & linear & linear & linear & linear & linear \\
        \midrule
        $z$-dropout & \emph{random-uniform}[0, 0.5] & 0.47 & 0.49 & 0.41 & 0.45\\
        \midrule
        learning rate optimizer & Adam  & Adam & Adam & Adam & Adam\\
        \midrule
        learning rate & \emph{loguniform-float}[1e-4, 1e-2] & 0.00081 & 0.00021 & 0.00024 & 0.0040 \\ 
        \midrule
        update background frequency & \emph{choice}[True, False] & False & False & False & False \\ 
        \midrule
        vocabulary size & 30000 & 30000 & 30000 & 30000 & 30000 \\ 
        \bottomrule
    \end{tabular}
    
    \vspace{3mm}\begin{tabular}{cc}
      \toprule
      \textbf{Dataset} & \textbf{\VAMPIRE~NPMI} \\ 
      \midrule
      \imdb & 0.131\\ 
      \midrule
      \ag & 0.224  \\ 
      \midrule
      \yahoo & 0.475 \\ 
      \midrule
      \hatespeech & 0.139 \\ 
      \bottomrule
    \end{tabular}
    
    \caption{\VAMPIRE~search space, best assignments, and associated performance on the four datasets we consider in this work. }
    \label{tab:vampire_hyperparameters_imdb}
\end{table*}

\subsection{VAMPIRE Search}
\label{sec:vampiresearch}

For the results presented in the paper, we varied the hyperparameters of \VAMPIRE~across a number of different dimensions, outlined in Table~\ref{tab:vampire_hyperparameters_imdb}.

\subsection{Classifier Search}
\label{sec:classifiersearch}

To choose a baseline classifier for which we experiment with all pretrained models, we performed a mix of manual tuning and random search over four basic classifiers: CNN, LSTM, Bag-of-Embeddings (i.e., Deep Averaging Networks), and Logistic Regression.

\begin{figure*}[t]
    \centering
    \hspace*{-1cm}
    \includegraphics[scale=0.35]{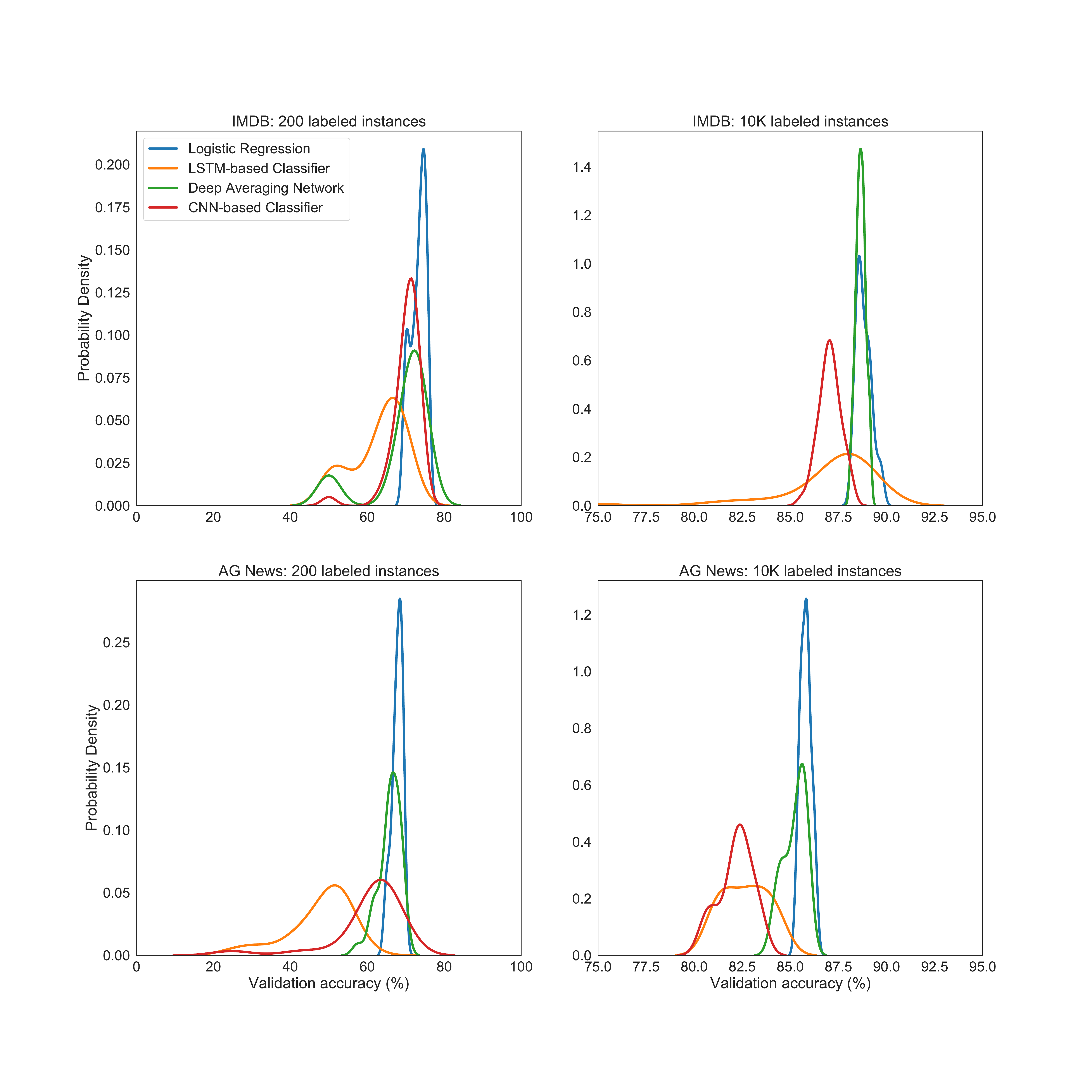}
    \caption{Probability densities of supervised classification accuracy in low-resource (200 labeled instances; left) and  high-resource (10K labeled instances; right) settings for \imdb~and \ag~datasets using randomly initialized trainable embeddings. Each search consists of 300 trials over 5 seeds and varying hyperparameters. We experiment with four different classifiers: Logistic Regression, LSTM-based classifier, Deep Averaging Network, and a CNN-based Classifier. We choose to use the Deep Averaging Network for all classifier baselines, due to its reliability, expressiveness, and low-maintenance. }
    \label{fig:density_plots}
\end{figure*}

Figure \ref{fig:density_plots} shows the distribution of validation accuracies using 200 and 10,000 labeled instances, respectively, for different classifiers on the \imdb~and \ag~datasets. Under the low-resource setting, we observe that logistic regression and DAN based classifiers tend to lead to more reliable validation accuracies. With enough compute, CNN-based classifiers tend to produce marginally higher validation accuracies, but the probability is mostly centered below those of the logistic regression and DAN classifiers. LSTM-based classifiers tend to have extremely high variance under the low-resource setting.  For this work, we choose to experiment with the DAN classifier, which comes with the richness of vector-based representations, along with the reliability that comes with having very few hyperparameters to tune.

\section{Results in the High Resource Setting}
\label{sec:appendix_results}

Table \ref{tab:resultshr} shows the results of all high-resource methods (along with \VAMPIRE) on all datasets, as we vary the amount of labeled data. As can be seen, training \elmo~\textit{only} on in-domain data results in similar or better performance to using an off-the-shelf \elmo~or \bert~model, \textit{without} fine-tuning it to in-domain data.

Except for one case in which it fails badly (\yahoo~with 200 labeled instances), fine-tuning \bert~to the target domain achieves the best performance in every setting. Though we performed a substantial hyperparameter search under this regime, we attribute the failure of fine-tuning \bert~under this setting to potential hyperparameter decisions which could be improved with further tuning. Other work has suggest that random initializations have a significant effect on the failure cases of BERT, pointing to the brittleness of fine-tuning \cite{Phang2018SentenceEO}.

The performance gap between fine-tuned \elmo~and frozen \elmo~in AG News corpus is much smaller than that of the other datasets, perhaps because the \elmo~model we used was pre-trained on the Billion Words Corpus, which is a news crawl. This dataset is also an example where frozen \elmo~tends to out-perform using \VAMPIRE. We attribute the strength of frozen, pretrained \elmo~under this setting as further evidence of the importance of in-domain data for effective semi-supervised text classification.

\begin{table*}[t!]
\small
\centering
\begin{tabular}{lrrrrr}
  \toprule
  \textbf{Dataset} & \textbf{Model} & \textbf{200} &  \textbf{500} &  \textbf{2500}  & \textbf{10000} \\
  \midrule
    \imdb & \elmo~(FR)  & 75.1 (1.4) & 80.3 (1.1) & 85.3 (0.1) & 87.3 (0.3)\\
    & \bert~(FR) & 81.5 (1.0) & 83.9 (0.4) & 86.8 (0.3) & 88.2 (0.3)\\
    & \elmo~(ID)  & 81.7 (1.3) & 84.5 (0.2) & 86.3 (0.4) & 88.0 (0.4) \\
    & \VAMPIRE  & 
      82.2 (2.0) & 84.5 (0.4) & 85.4 (0.4) & 87.1 (0.4)\\
    & \elmo~(FT) & 86.4 (0.6) & 87.9 (0.4) & 90.0 (0.4) & 91.6 (0.2) \\
    & \bert~(FT) & \textbf{88.1} (0.7) & \textbf{89.4} (0.7) & \textbf{91.4} (0.1) & \textbf{93.1} (0.1)\\
  \midrule
  \midrule
      \ag & \elmo~(FR)  & 84.5 (0.5) & 85.7 (0.5) & 88.3 (0.2) & 89.4 (0.3)\\
      & \bert~(FR)  & 84.6 (1.1) & 85.7 (0.7) & 88.0 (0.4) & 89.0 (0.3)\\
      & \elmo~(ID)  & 84.5 (0.6) & 85.8 (0.8) & 87.9 (0.2) & 89.2 (0.2)\\
      & \VAMPIRE  & 
      83.9 (0.6) & 84.5 (0.4) & 85.8 (0.2) & 87.7 (0.1)\\
      & \elmo~(FT) & 85.2 (0.5) & 86.6 (0.4) & 88.6 (0.2) & 89.5 (0.1)\\
      & \bert~(FT) & \textbf{87.1} (0.6) & \textbf{88.0} (0.4) & \textbf{90.1} (0.5) & \textbf{91.9} (0.1)\\
  \midrule
  \midrule
      \yahoo & \elmo~(FR) & 54.3 (1.6) & 64.2 (0.6) & 71.2 (1.3) & 74.1 (0.3) \\
      & \bert~(FR) & 57.0 (1.3) & 64.2 (0.5) & 70.0 (0.3) & 73.8 (0.2)\\
      & \elmo~(ID) & \textbf{60.9} (1.7) & 66.9 (0.9) & 72.8 (0.5) & 75.6 (0.1)\\
      & \VAMPIRE  & 
      59.9 (0.9) & 65.1 (0.3) & 69.8 (0.3) & 73.6 (0.2)\\
      & \elmo~(FT) & 60.5 (1.9) & 66.1 (0.7) & 71.7 (0.7) & 75.8 (0.3)\\
      & \bert~(FT) & 45.3 (7.5) & \textbf{69.2} (1.6) & \textbf{76.9} (0.6) & \textbf{81.0} (0.1) \\
  \midrule
  \midrule
      \hatespeech & \elmo~(FR)  & 70.5 (1.7) & 72.4 (0.9) & 76.0 (0.5) & 78.3 (0.2)\\
      & \bert~(FR) & 75.1 (0.6) & 76.3 (0.3) & 77.8 (0.4) & 79.0 (0.2) \\
      & \elmo~(ID)  & 73.3 (0.8) & 74.1 (0.8) & 77.2 (0.3) & 78.9 (0.2) \\
      & \VAMPIRE  & 
      74.1 (0.8) & 74.4 (0.5) & 76.2 (0.6) & 78.0 (0.3)\\
      & \elmo~(FT) & 73.9 (0.6) & 75.4 (0.4) & 78.1 (0.3) & 78.7 (0.1)\\
      & \bert~(FT) & \textbf{76.2} (1.8) & \textbf{78.3} (1.0) & \textbf{79.8} (0.4) & \textbf{80.2} (0.3)\\
      
  \bottomrule
\end{tabular}
\caption{Results in the high-resources setting. }

\label{tab:resultshr}
\end{table*}

\section{Further Details on NPMI vs. NLL as Stopping Criteria}
\label{sec:appendix_stopping}

\begin{figure}[t]
    \centering
    \hspace{-5.3mm}\includegraphics[scale=0.54]{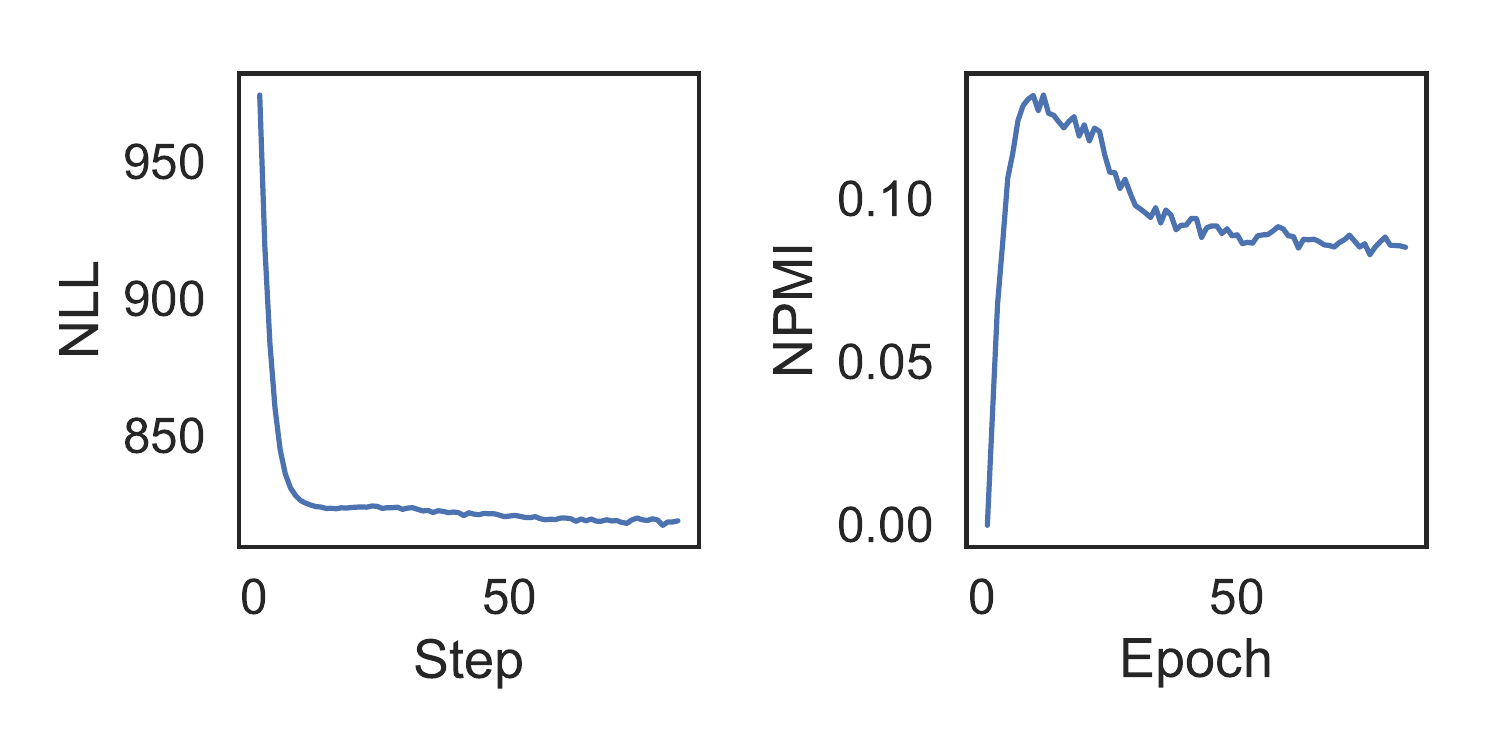}
    \caption{An example learning curve when training \VAMPIRE~on the \imdb~dataset. If trained for too long, we observe many cases in which NPMI (higher is better) degrades while NLL (lower is better) continues to decrease. To avoid selecting a model that has poor topic coherence, we recommend performing model selection with NPMI rather than NLL.}
    \label{fig:npmi_nll_appendix}
\end{figure}

In the main paper, we note that we have observed cases in which training \VAMPIRE~for too long results in NPMI degradation, while NLL continues to improve.  In Figure~\ref{fig:npmi_nll_appendix}, we display example learning curves that point to this phenomenon.

\section{Additional Learned Topics}
\label{sec:additionaltopics}

In Table~\ref{tab:add_topics} we display some additional topics learned by \VAMPIRE~on the \yahoo~dataset.

\begin{table*}[t!]
    \centering
    \begin{tabular}{cccc}

        \multicolumn{4}{c}{\yahoo}\\
        
        \toprule

        \textbf{Canine Care} & \textbf{Networking} & \textbf{Multiplayer Gaming} & \textbf{Harry Potter}  \\
        
        \midrule
        training & wireless & multiplayer & dumbledore  \\
         obedience & homepna & massively & longbottom \\ 
         schutzhund & network & rifle & hogwarts \\
         housebreaking & verizon & cheating & malfoy \\
         iliotibial & phone & quake & weasley \\
         crate & blackberry & warcraft & rubeus \\
         ligament & lan & runescape & philosopher \\
         orthopedic & telephone & socom & albus \\
         fracture & bluetooth & fortress & hufflepuffs \\
         gait & broadband & duel & trelawney  \\
        \bottomrule

    \end{tabular}
    
    \vspace{3mm}\begin{tabular}{cccc}

        \toprule

       \textbf{Nutrition} & \textbf{Baseball} & \textbf{Sexuality} & \textbf{Religion} \\
        
        \midrule
        nutritional & baseball & homophobia & islam \\
          obesity & sox & heterosexuality & jesus \\ 
         weight & yankees & orientation & isaiah  \\
         bodybuilding & rodriguez & transsexuality & semitism\\
         anorexia & gehrig & cultures & christian \\
         diet & cardinals & transgender & baptist\\
          malnutrition & astros & polyamory & jewish \\
          nervosa & babe & gay & prophet\\
         gastric & hitter & feminism & commandments\\
         watchers & sosa & societal & god\\
        \bottomrule

    \end{tabular}
    
    \caption{Example topics learned by \VAMPIRE~in the \yahoo~dataset.}
    \label{tab:add_topics}
\end{table*}

\end{document}